\documentclass{article}

\usepackage{PRIMEarxiv}

\usepackage[utf8]{inputenc} 
\usepackage[T1]{fontenc}    

\usepackage[numbers]{natbib}

\usepackage{hyperref}       
\usepackage{url}            
\usepackage{booktabs}       
\usepackage{amsfonts}       
\usepackage{apalike}
\usepackage{nicefrac}       
\usepackage{microtype}      
\usepackage{lipsum}
\usepackage{graphicx}

\usepackage{amsmath}

\usepackage{subcaption}

\newcommand{\vect}[1]{\mathbf{#1}}

\DeclareMathOperator*{\argmin}{argmin}

\usepackage{multirow}

\graphicspath{{figures/}}     

\title{Deep Learning Architectures for Code-Modulated Visual Evoked Potentials Detection
\thanks{
\textit{\underline{Corresponding Author:}} Hubert Cecotti. 
Department of Computer Science, College of Science and Mathematics, 
California State University, Fresno, CA, USA.  
Tel: (559) 278-2905.  
Email: hcecotti@csufresno.edu.
}
}

\author{
  Kiran Nair \\
  Department of Computer Science \\
  California State University, Fresno \\
  Fresno, CA, USA \\
  \texttt{kiranpnair8@mail.fresnostate.edu} \\
  \And
  Hubert Cecotti\thanks{Corresponding author.} \\
  Department of Computer Science \\
  California State University, Fresno \\
  Fresno, CA, USA \\
  \texttt{hcecotti@csufresno.edu} \\
}

\begin{document}
\maketitle

\begin{abstract}
Non-invasive Brain–Computer Interfaces (BCIs) based on Code-Modulated Visual Evoked Potentials (C-VEPs) require highly robust decoding methods to address temporal variability and session-dependent noise in EEG signals. This study proposes and evaluates several deep learning architectures, including convolutional neural networks (CNNs) for 63-bit m-sequence reconstruction and classification, and Siamese networks for similarity-based decoding, alongside canonical correlation analysis (CCA) baselines. EEG data were recorded from 13 healthy adults under single-target flicker stimulation. The proposed deep models significantly outperformed traditional approaches, with distance-based decoding using Earth Mover’s Distance (EMD) and constrained EMD showing greater robustness to latency variations than Euclidean and Mahalanobis metrics. Temporal data augmentation with small shifts further improved generalization across sessions. Among all models, the multi-class Siamese network achieved the best overall performance with an average accuracy of 96.89\%, demonstrating the potential of data-driven deep architectures for reliable, single-trial C-VEP decoding in adaptive non-invasive BCI systems.
\end{abstract}


\section{Introduction}
\label{section:introduction}

The rapid advancement of intelligent systems and neurotechnology has transformed how humans interact with machines, driving innovation across domains from healthcare to human augmentation. Among these emerging technologies, Brain-Computer Interfaces (BCIs) have established themselves as a revolutionary pathway enabling direct communication between the brain and external devices without requiring muscular output~\citep{gao2021interface}. BCIs are increasingly recognized as part of the broader evolution of intelligent systems, bridging neuroscience, signal processing, and machine learning. Initially designed to assist motor-impaired users following a spinal cord injury, or those who are unable to speak, such as those suffering from locked-in syndrome, BCIs have since expanded into diverse areas, including neurorehabilitation~\citep{mane2020bci,SIRVENTBLASCO20127908}, cognitive enhancement~\citep{de2018robust}, mental state monitoring, and entertainment applications~\citep{stimuli_nezamfar_2015}. Their ability to translate neural signals into actionable commands makes BCIs an integral component of the intelligent system ecosystem with significant potential for both clinical and non-clinical applications.

Despite notable progress, BCIs have yet to achieve the robustness, reliability, and user-friendliness required for widespread adoption. The main technical barriers include inconsistent performance across users and sessions, low signal-to-noise ratio (SNR), and sensitivity to artifacts such as eye blinks, muscle activity, and environmental interference~\citep{stateoftheart_guger_2011,dhanaselvam2023review}. Electroencephalography (EEG) remains the most widely used modality for non-invasive BCIs because of its affordability, safety, and portability. However, EEG signals are inherently noisy and spatially blurred, requiring effective preprocessing and robust decoding algorithms to achieve high accuracy. These challenges underscore the need for intelligent, data-driven approaches that can adapt to user variability and non-stationary signal conditions.

Among non-invasive paradigms, the Code-Modulated Visual Evoked Potential (c-VEP)\citep{alphaband_spler_2018,leveraging_he_2024,braincomputer_spler_2015,random_nagel_2016,MARTINEZCAGIGAL2023120815} has emerged as one of the most promising approaches for high-speed and reliable communication. In c-VEP systems, each target is encoded by a unique pseudorandom binary sequence, resulting in distinct time-locked EEG responses when the target is attended. Thanks to the excellent correlation properties of these codes, c-VEPs enable high classification accuracy and information transfer rates~\citep{towards_kiser_2025}. Recent advances have demonstrated impressive communication speeds exceeding 200~bits/min with minimal calibration~\citep{miao2024high}, establishing c-VEPs as strong candidates for real-world BCI deployment. Nevertheless, challenges persist, including subject-specific variability, temporal misalignment in neural responses, and the need for improved preprocessing to enhance spatial selectivity~\citep{martinez2021brain}. These issues motivate the development of advanced decoding methods and spatial enhancement strategies to ensure generalizable, session-independent performance.

This study proposes a systematic evaluation of both traditional and deep learning-based decoding strategies for c-VEP BCIs. Specifically, we compare classical feature extraction approaches such as Canonical Correlation Analysis (CCA) and correlation-based methods with Bayesian Linear Discriminant Analysis (BLDA) classifiers against modern neural architectures, including Convolutional Neural Networks (CNNs) and Siamese networks. Surface Laplacian filtering is integrated to enhance spatial resolution and suppress noise, while CNN outputs are evaluated using distance-based similarity metrics, including Euclidean, Mahalanobis, and Earth Mover’s Distance (EMD). Together, these components aim to provide a unified framework for robust, high-speed, and calibration-efficient c-VEP decoding.

The main contributions of this paper are summarized as follows:
\begin{enumerate}
    \item A comparative analysis of traditional correlation-based methods and deep learning architectures (CNN and Siamese networks) for c-VEP decoding.
    \item The development of a CNN framework that reconstructs 63-bit code patterns from EEG and classifies targets using distance-based metrics (Euclidean, Mahalanobis, and EMD).
    \item The analysis and comparison of single multi-class Siamese networks and multiple-classifier binary Siamese networks for evaluating cross-session generalization and temporal shift invariance in c-VEP decoding.
    \item An analysis of data augmentation and decision combination with local temporal shift in c-VEP decoding. 
\end{enumerate}

The remainder of this paper is organized as follows. Section~\ref{section:relatedwork} reviews prior work related to deep learning in BCIs, c-VEP-based paradigms, spatial filtering, and similarity learning. Section~\ref{section:methodology} describes the experimental setup for obtaining the signals and all the classifiers. Section~\ref{section:results} presents the performance of the different classifiers. Finally, the results are discussed in Section~\ref{section:discussion}, while Section~\ref{section:conclusion} summarizes the findings.
\section{Related Works}
\label{section:relatedwork}

Deep learning has profoundly influenced EEG-based BCI research by enabling data-driven feature extraction from raw neural signals. \cite{lawhern2018eegnet} introduced EEGNet, a compact CNN architecture that generalizes across paradigms such as P300, SSVEP, and motor imagery, setting a benchmark for lightweight yet robust models. \cite{Schirrmeister2017HBM} demonstrated that deep convolutional networks could learn interpretable spatiotemporal filters directly from EEG, outperforming traditional CSP-based approaches. More recent studies have refined these architectures with multi-branch 1D CNNs for frequency–time feature fusion~\cite{wrro199684}, as well as transformer-based attention mechanisms for dynamic feature representation~\citep{Wang2024EEGFMCNN}. Some studies demonstrated that deep learning methods, such as convolutional neural networks (CNNs), could effectively model electroencephalography (EEG) signals for P300-based detection, marking a milestone in end-to-end neural decoding~\citep{Cecotti2011TPAMI, ABINAYAA2026129674, YAN2025127859,ZHANG2018302}. Subsequent work emphasized the benefits of current-source density and surface Laplacian filtering for motor-imagery BCIs, demonstrating measurable gains in spatial resolution and classification performance~\citep{rathee2017current, SUN2024124144}. Such models not only improve accuracy but also reduce reliance on handcrafted features and extensive calibration, marking a paradigm shift toward generalizable, explainable neural decoding.

Convolutional Neural Networks (CNNs) are a class of artificial neural networks widely employed in deep learning for their ability to learn hierarchical feature representations directly from data~\citep{goodfellow2017deep}. Inspired by the organization of the human primary visual cortex~\citep{grill2004human}, CNNs consist of stacked convolutional, pooling, and fully connected layers that progressively extract increasingly abstract features from structured inputs such as images or multi-dimensional signals. Their effectiveness has been demonstrated across diverse domains, including computer vision~\citep{Cecotti2020ESWA, ciresan2012multi}, robotics~\citep{redmon2015real}, chemistry~\citep{ma2015deep}, and astronomy~\citep{dieleman2015rotation}. 


Within the family of visual evoked potentials, c-VEP-based BCIs have become a focal point for high-speed, reliable communication. Early concepts by Sutter~\citep{sutter1992brain} introduced pseudorandom code sequences to evoke distinct neural responses, later re-established by \cite{bin2009vep} using non-invasive EEG. \cite{wittevrongel2017code} enhanced c-VEP decoding through spatiotemporal beamforming, achieving over 170~bits/min, while \cite{miao2024high} demonstrated cross-subject transfer learning that achieved 250~bits/min with under one minute of calibration. \cite{martinez2021brain} provided a comprehensive review of current trends in code modulation, template correlation, and machine-learning-based decoding. Further studies have optimized code design~\citep{wei2016plos}, explored multi-target spellers~\citep{liu2018plos}, and proposed novel coding strategies such as chaotic or burst stimuli to enhance user comfort and signal distinctiveness~\citep{Wang2024EEGFMCNN}. \cite{volosyak2020towards} reported near-universal usability with 97.8\% accuracy across 86 subjects, mitigating the long-standing issue of BCI illiteracy.

In addition to code-level advances, signal preprocessing plays a critical role in improving EEG reliability~\citep{GAUR2018201,HAN2024121986}. Surface Laplacian (SL) and current source density (CSD) transformations have been shown to enhance spatial resolution by isolating local cortical activity while suppressing distant sources. Carvalhaes and de Barros~\cite{carvalhaes2015surface} formalized the theoretical foundations of SL methods, and \cite{mcfarland2015advantages} demonstrated their practical benefits for VEP-based BCIs. \cite{rathee2017current} indicated that SL filtering improved classification accuracy by 3–5\%, particularly under low SNR or sparse electrode conditions. These findings establish Laplacian filtering as an essential preprocessing step for modern EEG pipelines.

Recent developments~\citep{novel_gembler_2019,automatic_zarei_2022,evaluation_fodor_2024,optimization_behboodi_2020,ZHAO2024124145} have extended deep learning in BCIs beyond conventional classification into similarity-based learning. Siamese and twin-network architectures enable metric learning for EEG, allowing models to capture invariant relationships across sessions or users. \cite{shahtalebi2020siamese} introduced a Siamese CNN for motor imagery BCIs using one-vs-one scaling, while \cite{Li2022BSPC} employed Siamese networks for cross-subject EEG data augmentation. Multiscale Siamese CNNs~\citep{Jiang2021JNM} and recent attention-based twin networks~\citep{zhou2025siamesenetworkdualattention} have further advanced the extraction of discriminative EEG embeddings. Such metric-learning approaches are particularly relevant for c-VEP decoding\citep{multisymbol_ye_2022,svm_aminaka_2015,highspeed_grigoryan_2019,exploring_gembler_2020, ZHAO2021114961}, where robustness to latency shifts and inter-session variability is critical.

The literature reveals a convergence of innovations in c-VEP paradigms, spatial pre-processing, and deep metric learning. These works provide the scientific foundation and motivation for developing a unified decoding framework that combines spatial enhancement, correlation-based analysis, and deep neural architectures to achieve reliable, session-independent c-VEP classification.

\section{Methods}
\label{section:methodology}

\subsection{Experimental Protocol}
\label{sec:expro}

\subsubsection{Subjects}
The experimental protocol was designed to evaluate the performance of the proposed Code-Modulated Visual Evoked Potential (C-VEP)-based Brain–Computer Interface (BCI). The study was approved by the Committee for the Protection of Human Subjects (CPHS) - Institutional Review Board (IRB) at California State University, Fresno, and conducted in accordance with the principles of the Helsinki Declaration of 1975, as revised in 2000. 

\begin{figure}[t]
    \centering
    \includegraphics[width=0.95\linewidth]{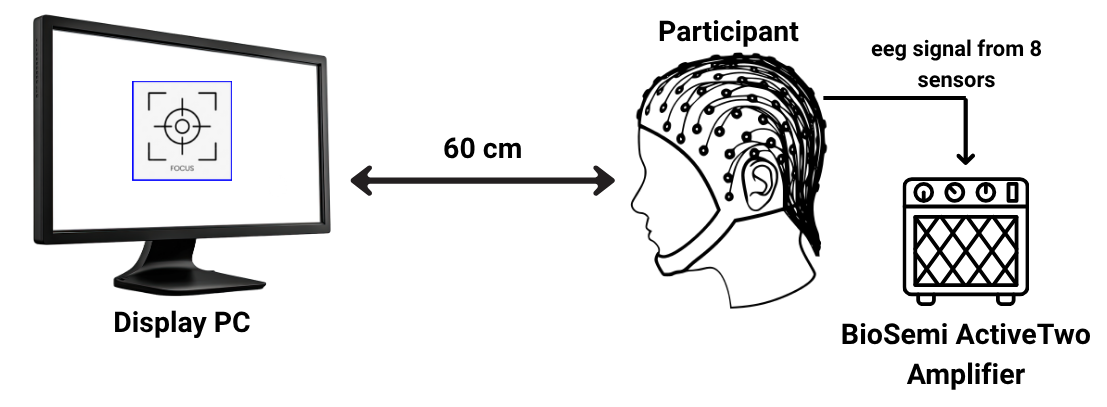}
    \caption{Schematic of the experimental setup. }
    \label{fig:exp_block_diagram}
\end{figure}

Thirteen healthy participants (10 males and 3 females; mean age = 23.54~$\pm$~6.74~years) voluntarily took part in the experiment. All subjects reported normal or corrected-to-normal vision and no prior history of neurological disorders. Each participant provided written informed consent before the experiment began. During data collection, participants were seated comfortably in a quiet, well-lit room, approximately 60~cm from the computer screen displaying the visual stimuli(see Fig.~\ref{fig:exp_block_diagram}).

\subsubsection{Signal Acquisition}
\begin{figure}[ht]
    \centering
    \includegraphics[width=0.45\linewidth]{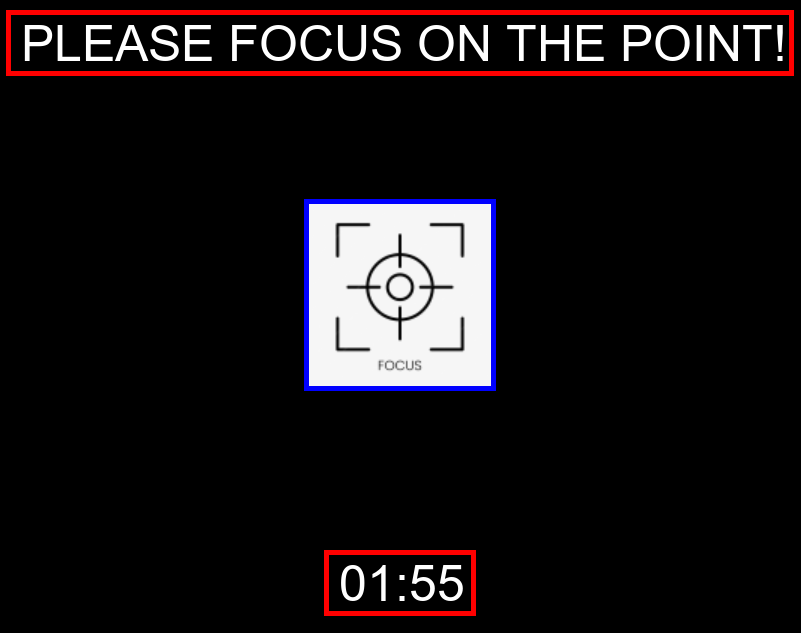} 
    \caption{Visual stimuli presented on the screen for recording the calibration data.}
    \label{fig:calibration}
\end{figure}

EEG signals were recorded using a BioSemi ActiveTwo amplifier system with a sampling rate of 512~Hz. A total of $N_s = 8$ active electrodes were positioned over the occipital and parietal regions according to the international 10–20 system\citep{HOMAN1987376,morley201610} at locations: O1, O2, Oz, Pz, P3, P4, PO7, PO8, as shown in Fig.~\ref{fig:calibration}. These electrode sites were selected for their known sensitivity to visual evoked potentials. The ground and reference electrodes were connected to the BioSemi common mode sense (CMS) and driven right leg (DRL) electrodes, respectively. A total of five sessions were recorded per participant ($N_{\mathrm{sessions}} = 5$).

\begin{figure}[!t]
    \centering
    \includegraphics[width=0.95\linewidth]{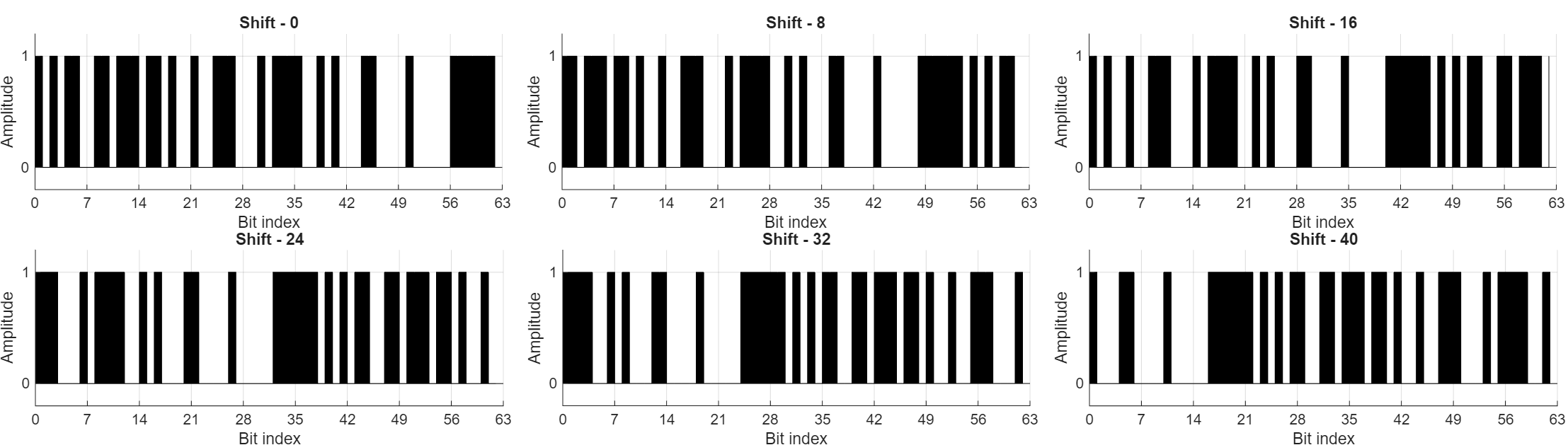}
    \caption{Binary representation of the $K$-bit $m$-sequence and its five circularly shifted variants (shifts of 8, 16, 24, 32, and 40 bits).}
    \label{fig:msequence_all}
\end{figure}

Given the screen refresh rate of 60~Hz and the $K$-bit stimulation sequence ($K = 63$), each visual epoch corresponded to 538 time points ($N_t = 538$), equivalent to approximately 1.05~s per trial. For each subject, $N_s$ electrodes were used, and six stimulus classes ($N_{\mathrm{classes}} = 6$) corresponding to circular code shifts of 0, 8, 16, 24, 32, and 40 bits (see Fig.~\ref{fig:msequence_all}) were considered.

\subsubsection{Experimental design}

\begin{figure}[!t]
    \centering
    \includegraphics[width=0.9\linewidth]{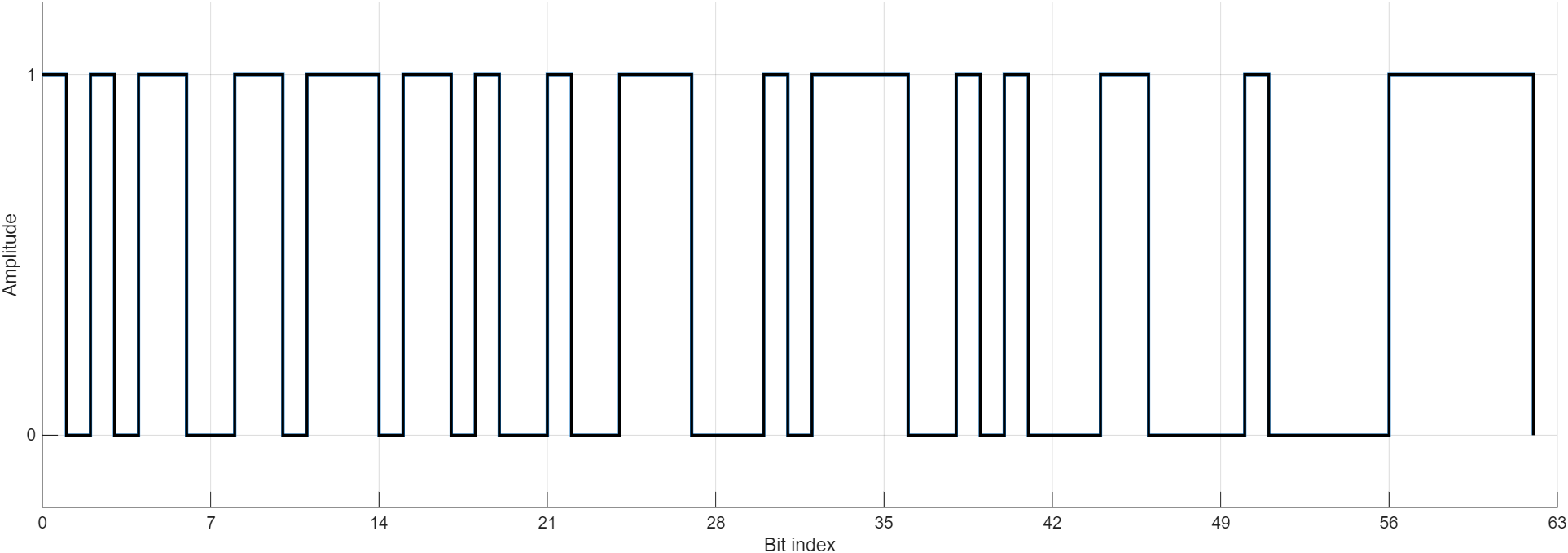}
    \caption{Binary representation of the $K$-bit $m$-sequence stimulus pattern used for C-VEP paradigm. 
    The sequence alternates between 0 and 1 according to the pseudorandom maximal-length code, forming the base temporal modulation for the visual stimulus.}
    \label{fig:msequence}
\end{figure}

\begin{figure}[ht]
    \centering
    \includegraphics[width=0.45\linewidth]{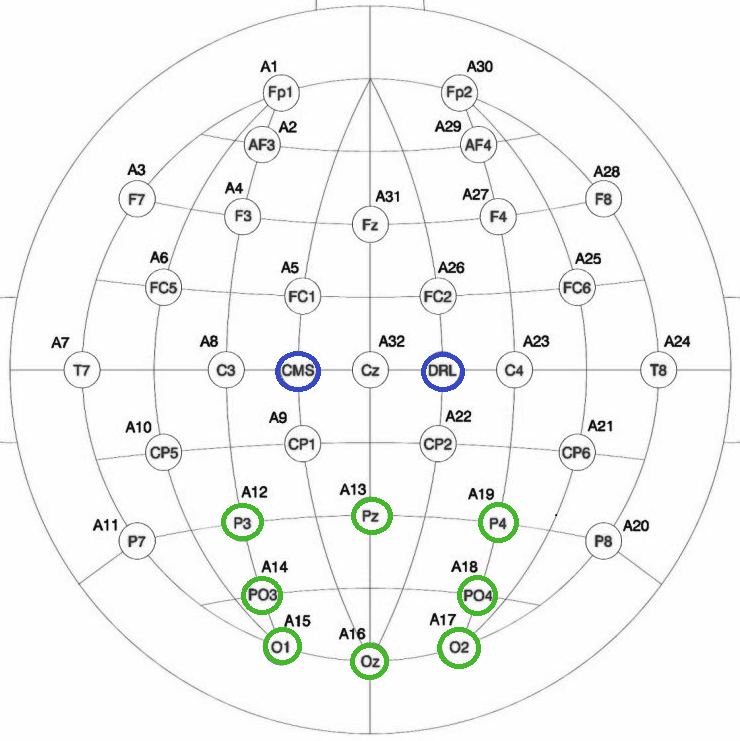}
    \caption{BioSemi 32+2 electrode layout. The highlighted electrodes(marked in blue(8 electrodes) and green(ground)) indicate those used for C-VEP signal acquisition.}
    \label{fig:biosemi}
\end{figure}

The calibration interface (see Fig.~\ref{fig:biosemi}) consisted of a single box flickering on a 27-inch screen (LG27GL83A, 1920 $\times$ 1080, 350 $cd/m^2$, 60~Hz refresh rate). The stimulus was 192 $\times$ 192, corresponding to about 6.22 $cd/m^2$. The flickering pattern was modulated using a K-bit m-sequence code~\citep{volosyak2020towards}, alternating between an on/off state (black/white image on the screen). The m-sequence code(see Fig.~\ref{fig:msequence}) used was: '10101100110111011010010011100010111100101000110\\
0001000001111110', where alternated between two states: a calibration image displayed for code~1 and a black screen for code~0. This distinct pattern ensured robust stimulation for generating C-VEP signals~\citep{inger2014potential}. The calibration interface, which displayed a flickering single box, was designed and implemented in MATLAB and Psychtoolbox to ensure precise timing and synchronization of the L-bit m-sequence visual stimuli~\citep{kleiner2007s}.


\subsection{Signal Processing}
\label{sec:Sigpro}

The raw EEG signals were first scaled to microvolts ($\mu V$) and preprocessed to enhance signal quality. Each channel was detrended to remove slow-varying baseline drifts and low-frequency fluctuations, ensuring a zero-mean baseline across all epochs. This step was essential for eliminating slow DC shifts and stabilizing the data for subsequent feature extraction and classification.
Following detrending, a bandpass filter with cutoff frequencies of $f_c^{\text{low}} = 0.5$~Hz and $f_c^{\text{high}} = 42.66$~Hz was applied to preserve the frequency range relevant to visual evoked potentials while attenuating muscle noise and low-frequency artifacts. The filtered signals were then segmented into stimulus-locked epochs of $N_t$ samples.

\subsubsection{Surface Laplacian Filtering with Great-Circle Distance}

Sensors are arranged on a spherical head model to account for scalp curvature. In this configuration, the distances between electrodes are calculated using the great-circle distance (orthodromic distance)~\citep{bullock2007great,porcu2016spatio,carter2002great}, ensuring that spatial relationships are preserved on the scalp surface. This step is fundamental for determining the influence of neighboring electrodes during signal processing. We denote by $d^{\mathrm{GCD}}_{ij}$ the great-circle distance between sensor $i$ and sensor $j$.

The great-circle distance between two points $(\phi_i,\lambda_i)$ and $(\phi_i,\lambda_i)$ on a sphere of radius $r$ is given by:
\begin{equation}
    d^{\mathrm{GCD}}_{ij} = r \cdot \Delta \sigma,
\end{equation}
where
\begin{equation}
\Delta \sigma = \mathrm{atan2}\!\Big(
\sqrt{ a_1 + a_2 },\, a_3 \Big),
\label{eq:vincenty}
\end{equation}
with
\begin{equation}
\begin{aligned}
a_1 &= (\cos\phi_j \sin\Delta\lambda)^2,\\
a_2 &= (\cos\phi_i \sin\phi_j - \sin\phi_i \cos\phi_j \cos\Delta\lambda)^2,\\
a_3 &= \sin\phi_i \sin\phi_j + \cos\phi_i \cos\phi_j \cos\Delta\lambda.
\end{aligned}
\end{equation}
With $\Delta\lambda = \lambda_j - \lambda_i$ being the difference in longitude between the two sensors. This is based on the Vincenty formula for great-circle distance. 

The weights $w_{ij}$ determine the influence of neighboring electrodes and are computed using the following equation:
\begin{equation}
  w_{ij} = \frac{1}{d^{\mathrm{GCD}}_{ij}}
\end{equation}

The Laplacian-filtered signal $S_i'$ is then estimated as:
\begin{equation}
    S_i' = S_i - \sum_{j \neq i} w_{ij} \cdot S_j
\end{equation}
where $S_i$ is the signal at the target electrode $i$ and $S_j$ is the signal from neighboring electrodes $1 \leq j \leq N_{s}$, $i \neq j$, where $N_s$ is the number of sensors. This transformation enhances local neural activity by reducing the effect of distant sources. The filter applies weights to electrodes within a defined radius of 1 (in normalized units). 

\subsection{Feature extraction}

\subsubsection{Canonical Correlation Analysis (CCA)}

Canonical Correlation Analysis (CCA) was employed to extract correlation-based features that quantify the relationship between the recorded EEG signals and reference templates corresponding to each visual stimulus shift\citep{bykhovskaya2025canonicalcorrelationanalysisreview, 10261825, articlecca}. The resulting canonical correlation coefficients $\rho=[\rho_1,\ldots,\rho_{n_f}]$ capture the linear associations between the EEG projections and stimulus templates across electrodes.

Because one of the eight EEG channels exhibited high redundancy or excessive noise, the effective rank of the data matrix was reduced to seven. Consequently, only seven canonical correlations were computed per trial, reflecting the number of independent spatial components. To maintain a uniform feature length across all samples, an additional zero value was appended, yielding consistent eight-dimensional vectors per shift. Concatenating these across all six code models produced a $1\times48$ feature vector for each example. The resulting CCA features captured spatially distinct stimulus-specific correlations while ensuring dimensional consistency for subsequent Bayesian Linear Discriminant Analysis (BLDA) classification.

\subsubsection{Correlation Coefficients}

Correlation-based feature extraction was employed to quantify the similarity between the recorded EEG responses and the expected m-sequence templates associated with each visual stimulus~\citep{schober2018correlation}. For each subject, $N_s$ electrodes were used, and $N_{classes}$ stimulus classes corresponding to circular code shifts of 0, 8, 16, 24, 32, and 40 bits were considered. Each class contained eight template signals, resulting in a total of $N_s \times N_s \times N_{classes} = 384$ correlation features per trial. 

Each data block comprised 114 trials($N_{examples} = 114$), each representing an L-bit stimulation period. After applying six circular shifts to represent the $N_{classes}$ classes, a total of 684 examples were generated per block ($N_{examples}$ per class). The extracted correlation features were organized into training and testing matrices, $X_{\text{train}}$ and $X_{\text{test}}$, each of dimension $684 \times 384$. These correlation coefficients served as discriminative representations of the neural response patterns corresponding to each visual stimulus, facilitating robust classification across subjects and sessions.

\subsubsection{Deep Learning Features}
\label{dl_features}

The early convolutional layers of the CNN were used to learn discriminative spatial–temporal representations from raw EEG signals and it consisted of 11 layers with approximately 8.6k learnable parameters. The network began with an input layer of size $N_s\times N_t\times1$, corresponding to the eight spatial channels and $N_t$ temporal samples. A spatial convolution layer with $N_s\times1$ kernels was applied to capture inter-channel dependencies and enhance localized spatial information. 

Two subsequent temporal convolutional blocks progressively extracted dynamic temporal patterns from the EEG. The first block used $1\times11$ kernels with 16 filters, followed by a ReLU activation, $1\times2$ max pooling, and a 20\% dropout for regularization. The second block employed $1\times14$ kernels with 32 filters, also followed by ReLU activation, $1\times2$ max pooling, and a 30\% dropout. These initial layers generated compact, informative spatial–temporal feature maps that were later used for classification and similarity learning.

\subsection{Classifiers}

\subsubsection{Bayesian Linear Discriminant Analysis (BLDA)}


BLDA extends traditional Linear Discriminant Analysis by introducing a Bayesian regularization framework that automatically balances model complexity and generalization performance\citep{comparing_gokcen_2002,bayesian_shen_2019,feature_huang_2007}. A multiclass BLDA approach was implemented using a one-vs-rest strategy, where an independent binary classifier was trained for each class. Each BLDA model estimates class-conditional distributions and computes posterior probabilities based on learned discriminant functions, providing robust decision boundaries even in high-dimensional and small-sample settings. During testing, the class with the highest posterior score was selected as the predicted label, while normalized class probabilities served as confidence estimates. 


\subsubsection{Distances}

The first CNN architecture was designed to reconstruct the 63-bit stimulus code directly from single-trial EEG recordings. Each input trial consisted of an $N_s\times N_t$ segment representing eight spatial EEG channels across $N_t$ temporal samples. The network output was a $K$-dimensional vector corresponding to the binary stimulation sequence used in the c-VEP paradigm. A sigmoid activation was applied to each output neuron to produce continuous values in the range [0, 1], enabling bitwise decoding and distance-based evaluation against the ground-truth 6$K$-bit templates.

The network was trained using the Adam optimizer with an initial learning rate of $1\times10^{-3}$ and an $L_2$ regularization factor of $1\times10^{-4}$ to prevent overfitting. Training was performed for up to 40 epochs with a mini-batch size of 64, and the data were shuffled at each epoch to improve generalization. Validation was conducted after each epoch using a held-out set \((X_{vdl}, Y_{vdl})\), and the root mean square error (RMSE) was monitored as the primary performance metric.

The network consisted of 2 more layers added to ~\ref{dl_features} with  127.1k added learnable parameters. The first few layers as mentioned in~\ref{dl_features} (as in Fig.~\ref{fig:cnnetwork})produced compact and informative spatial–temporal feature maps. These feature maps were flattened and passed through a fully connected layer with $K$ neurons, one corresponding to each bit in the stimulation code. Finally, a sigmoid activation layer generated the output vector of size $K\times1$, representing the reconstructed stimulus sequence.

\paragraph{Euclidean Distance Evaluation}

\label{sec:euclidean}

To evaluate network performance, a distance-based decoding approach was employed, comparing the CNN-reconstructed bit sequence with the reference code templates for each visual stimulus. For each trial \(i\), the network outputs a continuous-valued vector \(\hat{\vect{y}}_i \in [0,1]^K\), where \(K = 63\) represents the number of bits in the c-VEP stimulation code. This vector can be interpreted as the predicted probability or confidence associated with each bit position.

The Euclidean distance between the reconstructed output \(\bar{\vect{y}}_i\) (thresholded or mean-centered version of \(\hat{\vect{y}}_i\)) and each reference code template \(\vect{c}^{(N_{classes})}\) was computed as
\begin{equation}
d_{i,N_{classes}}^{(2)} = \bigl\|\, \bar{\vect{y}}_i - \vect{c}^{(N_{classes})} \,\bigr\|_2 
= \sqrt{\sum_{k=1}^{K}\!\bigl(\bar{y}_{i,k} - c^{(N_{classes})}_b\bigr)^2 } .
\end{equation}
Each distance \(d_{i,N_{classes}}^{(2)}\) quantifies the dissimilarity between the reconstructed sequence and the expected codeword for $N_{classes}$. The predicted class for trial \(i\) is determined as the one with the minimum Euclidean distance to the output vector:
\begin{equation}
N_{classes_i}^\star = \argmin_{N_c=1,\dots,N_{classes}} d_{i,N_c}^{(2)} 
= \argmin_{N_c} \sum_{k=1}^{K}\!\bigl(\bar{y}_{i,k} - c^{(N_c)}_k\bigr)^2 ,
\end{equation}
where the square root is omitted without affecting the decision outcome. 

This metric provides an intuitive, computationally efficient means of classifying reconstructed bit patterns, as smaller Euclidean distances indicate greater similarity between the predicted output and the true stimulus code.

\paragraph{Mahalanobis Distance Evaluation.}

The Mahalanobis distance was employed to account for feature correlations and varying variance among the reconstructed bits. This approach provides a more statistically informed similarity measure between the reconstructed output and the reference code templates. For each trial \(i\), the CNN output vector \(\hat{\vect{y}}_i \in [0,1]^K\) was compared against each class codeword \(\vect{c}^{(N_c)}\), using a shared covariance matrix estimated from the reconstructed outputs across all samples. The covariance matrix \(\boldsymbol{\Sigma}\) was regularized to ensure numerical stability via a shrinkage factor \(\lambda = 0.1\), forming
\begin{equation}
\boldsymbol{\Sigma}_{\text{reg}} = (1-\lambda)\boldsymbol{\Sigma} + \lambda\,\alpha\,\mathbf{I},
\end{equation}
where \(\alpha\) denotes the mean of the diagonal elements of \(\boldsymbol{\Sigma}\) and \(\mathbf{I}\) is the identity matrix. The Mahalanobis distance between the reconstructed output and each code template was then computed as
\begin{equation}
d_{i,N_c}^{(M)} = \sqrt{(\bar{\vect{y}}_i - \vect{c}^{(N_c)})^{\!\top} \boldsymbol{\Sigma}_{\text{reg}}^{-1} (\bar{\vect{y}}_i - \vect{c}^{(N_c)}) }.
\end{equation}
The predicted class label was assigned according to the minimum Mahalanobis distance:
\begin{equation}
N_{classes_i}^\star = \argmin_{N_c=1,\dots,N_{classes}} d_{i,N_c}^{(M)}.
\end{equation}

\paragraph{Earth Mover’s Distance (EMD) Evaluation.}

The EMD was employed as a distribution-based metric. Unlike Euclidean or Mahalanobis distances, which measure pointwise differences, EMD quantifies the minimal cumulative “cost” required to transform one probability distribution into another, providing a more perceptually meaningful measure of dissimilarity for structured signals such as the $K$-bit c-VEP codes. For each trial \(i\), the CNN output \(\hat{\vect{y}}_i \in [0,1]^K\) and each reference template \(\vect{c}^{(N_c)}\) were first normalized to form discrete probability mass functions (PMFs),
\begin{equation}
\vect{p}_i = \frac{\max(0,\,\hat{\vect{y}}_i)}{\sum_b \max(0,\,\hat{y}_{i,b})}, 
\qquad
\vect{q}_{N_c} = \frac{\vect{c}^{(N_c)}}{\sum_b c^{(N_c)}_b},
\end{equation}
ensuring that \(\sum_b p_{i,b} = \sum_b q_{N_c,b} = 1\). The one-dimensional EMD between \(\vect{p}_i\) and each class template \(\vect{q}_{N_c}\) was computed using the cumulative distribution formulation,
\begin{equation}
\begin{aligned}
d_{i,N_c}^{(\mathrm{EMD})} 
&= \frac{1}{K} \sum_{k=1}^{K} 
\bigl|\, \mathrm{CDF}_{p_i}(k) - \mathrm{CDF}_{q_{N_c}}(k) \bigr|, \\
&= \mathrm{mean}\!\left( \big| \mathrm{cumsum}(\mathbf{p}_i - \mathbf{q}_{N_c}) \big| \right).
\end{aligned}
\end{equation}

The predicted class was determined by selecting the template with the minimum EMD value:
\begin{equation}
N_{classes}{_i}^\star = \argmin_{N_c=1,\dots,N_{classes}} d_{i,N_c}^{(\text{EMD})}.
\end{equation}
By comparing cumulative distributions rather than individual bits, EMD captures global structural deviations between reconstructed and reference codes.

\paragraph{Constrained-EMD Evaluation.}

Contrained-EMD introduces a movement radius \(R\), restricting the transport of probability mass between bit positions to a local neighborhood, thereby emphasizing short-range structural correspondence in the reconstructed codes. For each trial \(i\), the predicted vector \(\hat{\vect{y}}_i \in [0,1]^K\) was normalized into a probability mass function (PMF), and compared against the set of reference code templates \(\vect{c}^{(N_c)}\), each also normalized to form \(\vect{q}_{N_c}\). The resulting optimization seeks the minimum transport cost required to transform \(\vect{p}_i\) into \(\vect{q}_{N_c}\), under the constraint that transport is only allowed between positions whose bit indices differ by at most \(R\):
\begin{equation}
\begin{aligned}   
\min_{\mathbf{X} \ge 0} \;\; \sum_{k,j} |k-j|\,X_{k,j}
\quad
\text{s.t.}\quad
\sum_j X_{k,j} = p_{i,k},\\ \quad
\sum_k X_{k,j} = q_{k,j}, \quad
X_{k,j} = 0 \;\text{if}\; |k-j| > R .
\end{aligned}
\end{equation}
Here, \(\mathbf{X}\) denotes the transport flow matrix between the distributions \(\vect{p}_i\) and \(\vect{q}_{N_c}\). The problem was solved using a linear programming formulation based on the dual-simplex algorithm with equality constraints enforcing mass conservation. When \(R\) is large (\(R \ge K-1\)), the formulation naturally reduces to the unconstrained 1-D EMD case based on cumulative distribution differences.

The constrained-EMD distance for each class \(N_c\) was computed as the minimal transport cost normalized by the total flow, and the predicted label was assigned to the class with the smallest constrained distance:
\begin{equation}
N_{classes_i}^\star = \argmin_{N_c=1,\dots,N_{classes}} d_{i,N_c}^{(\text{CEMD})}.
\end{equation}

\subsubsection{CNN}

A CNN-based $N_{classes}$ classification architecture was designed to classify EEG responses directly into six distinct classes corresponding to the six circular shifts (0, 8, 16, 24, 32, and 40 bits) of the $K$-bit m-sequence used in the c-VEP stimulation.

\begin{figure}[!ht]
    \centering
    \includegraphics[width=0.95\linewidth]{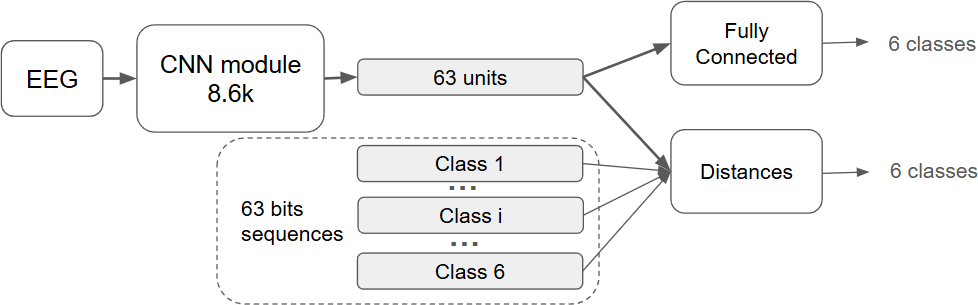}
    \caption{Schematic representation of the CNN models used for direct classification of the responses and to be used with distances.}
    \label{fig:cnnetwork}
\end{figure}

The network was trained using the Adam optimizer with an initial learning rate of $1\times10^{-3}$ and an $L_2$ regularization coefficient of $1\times10^{-4}$ to mitigate overfitting. The model was trained for 40 epochs with a mini-batch size of 64, and data was shuffled at each epoch to ensure robust learning. The training employed categorical cross-entropy as the loss function and accuracy as the primary evaluation metric. The network comprised 3 more layers added to \ref{dl_features} with approximately 12.1k added learnable parameters. The feature maps from \ref{dl_features} were flattened into a one-dimensional representation and passed to a fully connected layer with six neurons. The Softmax activation at the output produced a probability distribution over $N_{classes}$, enabling direct classification of the attended stimulus (see Fig.~\ref{fig:cnnetwork}).

The trained network generated predictions for the test set via forward propagation, producing a $N_{classes}\times N_{examples}$ matrix of class probabilities for each trial. The predicted class label $\hat{y}$ for each trial was obtained as the index of the highest confidence value. 

\subsubsection{Siamese Networks}

\paragraph{Single multi-class Siamese (Twin) Network}
\label{sec:cnn_siamese}
The third architecture is a convolutional Siamese (twin) network configuration designed to learn a similarity metric between pairs of EEG trials from different classes. Unlike previous architectures that performed classification directly, this model learns a shared embedding space in which trials from the same stimulus class produce similar representations, while trials from different classes are mapped farther apart. Each input pair consisted of two single-trial EEG segments of size $N_s\times N_t$, processed through identical CNN Modules with shared weights. The Siamese model output was a single probability representing the likelihood that the two inputs originated from the same c-VEP code shift.

The subnetwork consisted of one additional layer to~\ref{dl_features}, with approximately 127.1k additional parameters followed by the same spatial–temporal design as the previous CNNs. The feature maps from~\ref{dl_features} were flattened and projected into a $K$-dimensional embedding space, which served as the compact feature representation for each EEG trial.

\begin{figure}[!ht]
    \centering
    \includegraphics[width=0.95\linewidth]{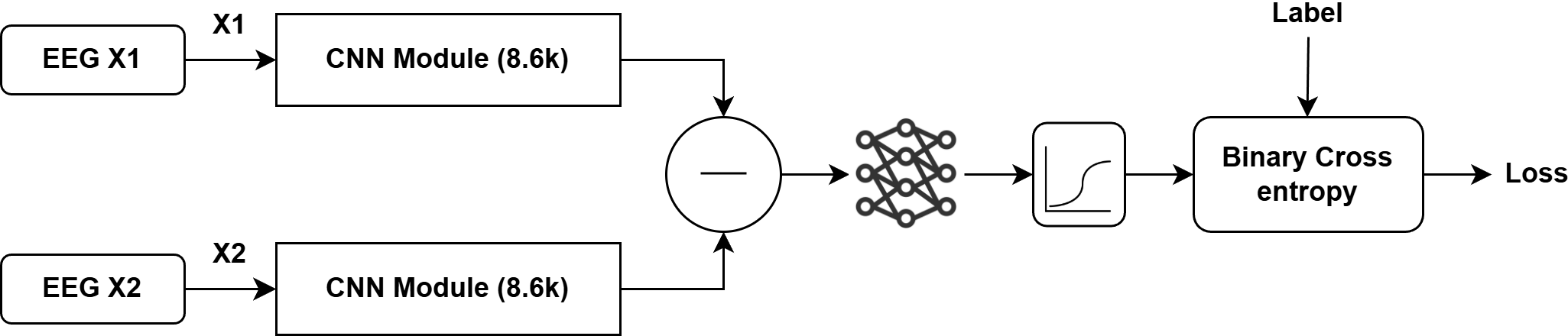}
    \caption{Schematic representation of the Siamese (Twin) CNN architecture. 
    The model consists of two identical CNN Modules that process paired EEG inputs ($X_1$, $X_2$) with shared weights. }
    \label{fig:twinnetwork}
\end{figure}

The similarity estimation between the two inputs was computed as the absolute difference between their embeddings, followed by a fully connected layer and a sigmoid activation producing a scalar similarity probability (see Fig~\ref{fig:twinnetwork}). The network was trained using the binary cross-entropy loss between predicted similarity scores and binary pair labels (1 for matching class, 0 otherwise). Training was conducted for 30 epochs with a batch size of 256 randomly balanced pairs per iteration. The Adam optimizer was used with a learning rate of $1\times10^{-3}$, and the loss was backpropagated through both subnetworks simultaneously to update the shared parameters.

\paragraph{Multi-classifier binary Siamese Network}
\label{sec:cnn_multisiamese}

The final configuration extended the convolutional Siamese framework to a multi-classifier setting, where separate Siamese models were trained for each of the $N_{classes}$. While the underlying architecture remained identical to the Single multi-class Siamese network described in Section~\ref{sec:cnn_siamese}, the training procedure was modified to produce class-specific similarity models that collectively enable multi-class discrimination. Each individual model was trained to distinguish between matching and non-matching EEG trial pairs associated with a single target code shift, effectively learning a dedicated embedding subspace for that particular class.

For each class \(k \in \{1,\dots,N_{classes}\}\), a Siamese model was trained using EEG trial pairs drawn from within and outside the class. The training data consisted of balanced positive (same-class) and negative (different-class) pairs, ensuring robust similarity learning. Training was conducted for 30 epochs with a batch size of 256 pairs, using the Adam optimizer with a learning rate of $1\times10^{-3}$.

During inference, the trained set of six Siamese models was evaluated independently. For each test pair, the similarity probability was computed for all $N_{classes}$, each yielding a binary classification output. 


\subsection{Performance Analysis}

The performance of the proposed methods was evaluated using a five-fold cross-validation scheme based on $N_{sessions}$ per subject. In each fold, one recording consisting of $N_{examples}$ was used for testing, while the remaining $N_{sessions} - 1$ ($4\times  N_{examples}$)  were used for training. The results reported in the subsequent tables represent the mean accuracy and standard deviation computed across the five folds for each subject.

\section{Results}
\label{section:results}

\begin{figure*}
    \centering
    \includegraphics[width=\textwidth]{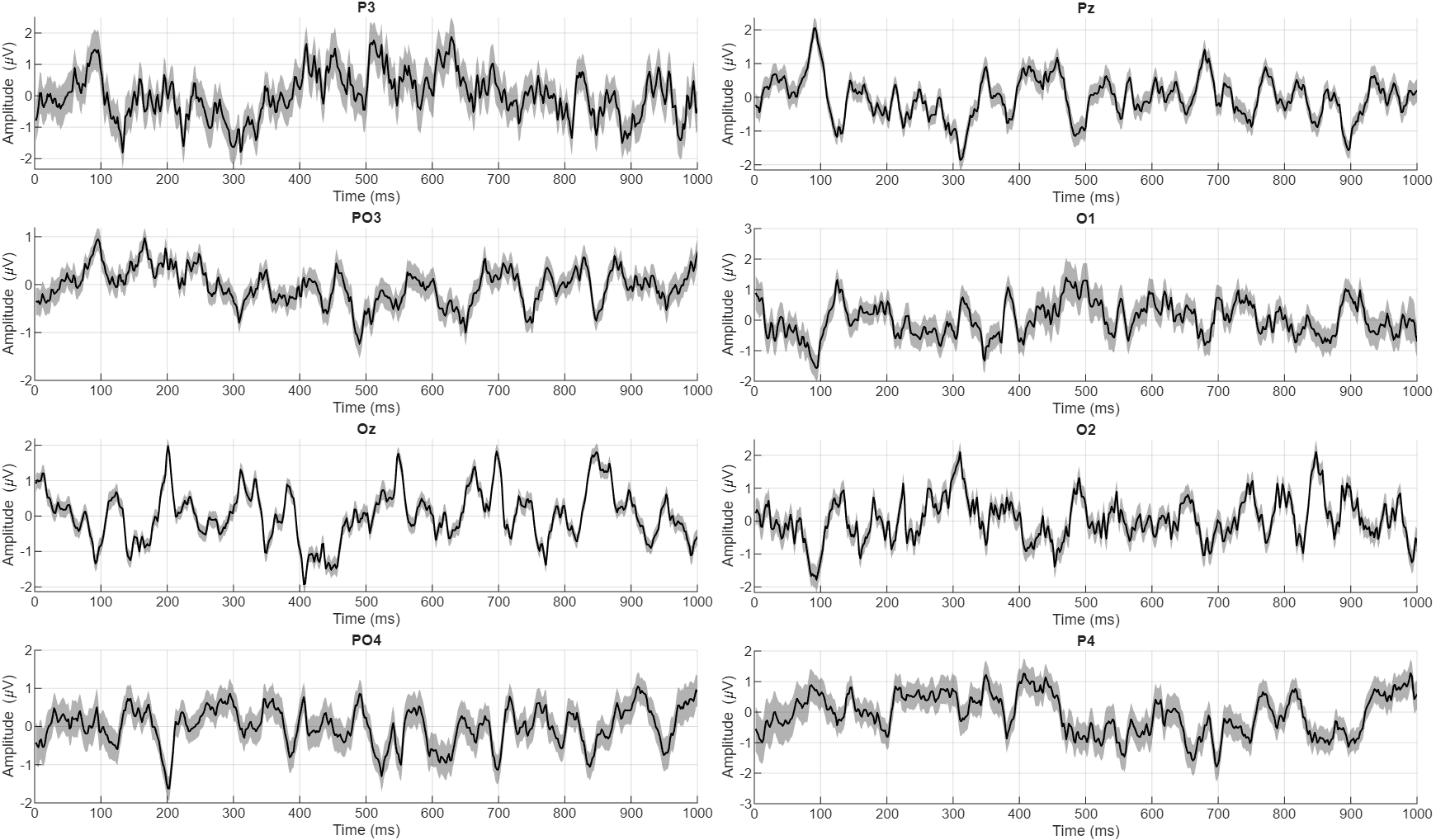} 
    \caption{Time-domain plots of EEG signals from eight selected electrodes.}
    \label{fig:time_domain_plot}
\end{figure*}

\begin{figure*}
    \centering
    \includegraphics[width=\textwidth]{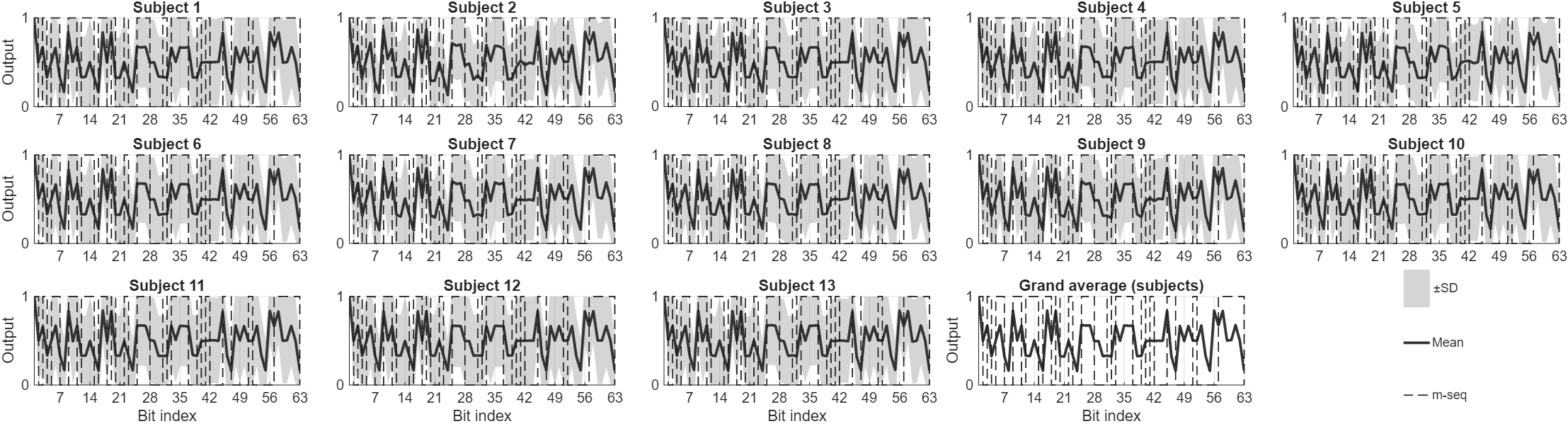}
    \caption{Grand-average of the CNN’s $K$-bit outputs across all subjects. }
    \label{fig:grandavg_63bit_per_subject}
\end{figure*}

\subsection{Grand Average Plots}

To visualize the consistency and reliability of neural responses across participants and sessions, grand-average analyses were performed. Fig.~\ref{fig:time_domain_plot} depicts time-domain plots of EEG signals from eight selected electrodes showing the mean response across trials . The shaded envelopes represent the standard error across participants, illustrating the trial-to-trial variability and consistency of neural responses. These plots highlight the stability of the signal features used for template-based classification in the C-VEP BCI system.

Fig~\ref{fig:grandavg_63bit_per_subject} provides an overview of the signals and the reconstructed $K$-bit outputs obtained from the proposed CNN-based model. Each subplot corresponds to a single participant and displays the CNN’s $K$-bit output, averaged across all trials and sessions. The dark gray line indicates the mean reconstructed signal, the shaded gray area represents the $\pm$SD range, and the dashed trace denotes the original $K$-bit $m$-sequence used for stimulation. The final panel shows the overall grand average and variability across subjects. The averaged waveforms highlight the network’s ability to capture the temporal structure of the stimulation code, offering insight into the spatial–temporal patterns that underpin classification performance.

\subsection{Classifier Performance}
\label{sec:classifier_performance}

The classification accuracy across 13 participants using the four methodological categories, correlation-based, canonical correlation-based, CNN-based, and Siamese-based networks, is summarized in Table~\ref{tab:corr_cca_blda}, Table~\ref{tab:cnn63bit_results}, Table~\ref{tab:cnn6class_results}, and Table~\ref{tab:siamese_results}, respectively. Among the conventional baselines, Corr+BLDA achieved a mean accuracy of $81.64 \pm 14.02\%$, outperforming CCA+BLDA ($62.59 \pm 16.57\%$), confirming the advantage of correlation-driven feature representation with Bayesian regularization for C-VEP decoding.  

In contrast, all deep learning models exhibited substantially higher performance, demonstrating the efficacy of end-to-end feature learning. The CNN-based $K$-bit models achieved accuracies of $93.86 \pm 6.16\%$ (Euclidean), $92.38 \pm 8.34\%$ (Mahalanobis), $93.88 \pm 6.38\%$ (EMD), and $91.30 \pm 7.99\%$ (Constrained-EMD). The CNN+$N_{classes}$ classifier reached $91.42 \pm 7.93\%$, indicating strong generalization despite reduced output dimensionality. The Siamese networks yielded $87.29 \pm 8.00\%$ (Single) and $96.89 \pm 2.74\%$ (multiplewith the latter achieving the highest overall mean accuracy overall. These results highlight that deep architectures, particularly distance-based CNNs and the Siamese multiple-classifier variant, significantly outperform traditional BLDA methods.

\begin{table}[!ht]
\centering
\caption{Classification accuracy (\%) comparison between feature extraction using Correlation coefficients and classification using BLDA and feature extraction using CCA and classification using BLDA across subjects.}
\label{tab:corr_cca_blda}
\begin{tabular}{c|c|c}
\hline
\textbf{Subject} & \textbf{Corrcoef+BLDA} & \textbf{CCA+BLDA} \\ 
\hline
1  & 88.72 $\pm$ 6.83 & 59.39 $\pm$ 6.96 \\
2  & 63.36 $\pm$ 25.35 & 59.79 $\pm$ 6.73 \\
3  & 99.24 $\pm$ 0.38 & 88.68 $\pm$ 4.23 \\
4  & 91.94 $\pm$ 4.00 & 79.14 $\pm$ 4.48 \\
5  & 50.21 $\pm$ 9.38 & 66.42 $\pm$ 9.54 \\
6  & 90.11 $\pm$ 2.19 & 67.92 $\pm$ 6.93 \\
7  & 73.42 $\pm$ 17.38 & 45.00 $\pm$ 11.20 \\
8  & 77.89 $\pm$ 3.97 & 32.14 $\pm$ 4.29 \\
9  & 71.12 $\pm$ 28.93 & 50.35 $\pm$ 17.49 \\
10 & 84.66 $\pm$ 4.94 & 49.66 $\pm$ 7.11 \\
11 & 87.81 $\pm$ 3.13 & 53.89 $\pm$ 3.83 \\
12 & 97.17 $\pm$ 1.40 & 80.90 $\pm$ 5.64 \\
13 & 85.65 $\pm$ 6.30 & 83.33 $\pm$ 3.26 \\
\hline
\textbf{Mean} & \textbf{81.64 $\pm$ 8.78} & \textbf{62.59 $\pm$ 7.05} \\
\textbf{SD}   & \textbf{14.02 $\pm$ 9.25} & \textbf{16.57 $\pm$ 3.89} \\
\hline
\end{tabular}
\end{table}

\begin{table*}[!ht]
\centering
\caption{Performance comparison (\%) of the $K$-bit reconstruction CNN using different distance evaluation metrics.}
\label{tab:cnn63bit_results}
\begin{tabular}{c|cccc}
\hline
\textbf{Subject} & 
\textbf{Euclidean Evaluation} & 
\textbf{Mahalanobis Evaluation} & 
\textbf{EMD Evaluation} & 
\textbf{Constrained EMD Evaluation} \\ 
\hline
1  & 99.39 $\pm$ 0.72 & 98.86 $\pm$ 1.14 & 98.04 $\pm$ 1.29 & 97.60 $\pm$ 2.14 \\
2  & 83.21 $\pm$ 29.22 & 71.39 $\pm$ 32.64 & 80.88 $\pm$ 27.26 & 70.08 $\pm$ 29.87 \\
3  & 99.68 $\pm$ 0.39 & 99.62 $\pm$ 0.43 & 99.44 $\pm$ 0.36 & 99.27 $\pm$ 0.36 \\
4  & 98.48 $\pm$ 1.05 & 96.90 $\pm$ 2.05 & 95.91 $\pm$ 1.59 & 95.35 $\pm$ 2.52 \\
5  & 86.42 $\pm$ 6.69 & 87.03 $\pm$ 7.24 & 83.75 $\pm$ 7.38 & 85.25 $\pm$ 6.66 \\
6  & 96.98 $\pm$ 0.97 & 95.99 $\pm$ 1.89 & 95.11 $\pm$ 1.11 & 94.73 $\pm$ 1.78 \\
7  & 84.95 $\pm$ 21.99 & 84.83 $\pm$ 23.63 & 83.96 $\pm$ 20.97 & 84.46 $\pm$ 20.95 \\
8  & 94.30 $\pm$ 3.06 & 93.54 $\pm$ 1.89 & 91.26 $\pm$ 3.54 & 91.11 $\pm$ 2.66 \\
9  & 86.81 $\pm$ 22.04 & 85.09 $\pm$ 24.34 & 93.13 $\pm$ 25.12 & 83.57 $\pm$ 23.92 \\
10 & 95.06 $\pm$ 3.55 & 94.71 $\pm$ 3.17 & 93.45 $\pm$ 3.42 & 93.25 $\pm$ 3.30 \\
11 & 97.57 $\pm$ 0.80 & 97.22 $\pm$ 0.48 & 96.40 $\pm$ 0.91 & 96.59 $\pm$ 0.62 \\
12 & 99.09 $\pm$ 0.76 & 99.04 $\pm$ 0.76 & 98.77 $\pm$ 0.82 & 98.65 $\pm$ 0.91 \\
13 & 98.30 $\pm$ 0.77 & 98.22 $\pm$ 0.64 & 97.16 $\pm$ 1.12 & 97.37 $\pm$ 1.10 \\
\hline
\textbf{Mean} & \textbf{93.86 $\pm$ 7.08} & \textbf{92.38 $\pm$ 7.57} & \textbf{93.88 $\pm$ 7.84} & \textbf{91.30 $\pm$ 7.55} \\
\textbf{SD}   & \textbf{6.16 $\pm$ 10.18} & \textbf{8.34 $\pm$ 10.24} & \textbf{6.38 $\pm$ 8.87} & \textbf{7.99 $\pm$ 8.98} \\
\hline
\end{tabular}
\end{table*}

\begin{table}[!ht]
\centering
\caption{Classification accuracy (\%) of the $N_{classes}$ CNN across all subjects.}
\label{tab:cnn6class_results}
\begin{tabular}{c|c}
\hline
\textbf{Subject} & \textbf{CNN + $N_{classes}$} \\
\hline
1  & 97.43 $\pm$ 1.75 \\
2  & 77.17 $\pm$ 30.88 \\
3  & 99.30 $\pm$ 0.30 \\
4  & 95.58 $\pm$ 2.28 \\
5  & 79.70 $\pm$ 10.14 \\
6  & 94.79 $\pm$ 1.76 \\
7  & 80.37 $\pm$ 23.20 \\
8  & 92.49 $\pm$ 2.80 \\
9  & 84.82 $\pm$ 24.26 \\
10 & 94.42 $\pm$ 3.63 \\
11 & 97.19 $\pm$ 0.73 \\
12 & 98.51 $\pm$ 0.82 \\
13 & 96.73 $\pm$ 1.22 \\
\hline
\textbf{Mean} & \textbf{91.42 $\pm$ 7.98} \\
\textbf{SD}   & \textbf{7.93 $\pm$ 10.76} \\
\hline
\end{tabular}
\end{table}

\begin{table}[!ht]
\centering
\caption{Performance comparison of Siamese network configurations: Single multi-class Siamese vs. Multiple-classifier binary Siamese.}
\label{tab:siamese_results}
\begin{tabular}{c|cc}
\hline
\textbf{Subject} & \textbf{Multi-class Siamese} & \textbf{Multi-classifier Siamese} \\ 
\hline
1  & 92.67 $\pm$ 3.60 & 98.39 $\pm$ 0.93 \\
2  & 75.64 $\pm$ 15.17 & 92.73 $\pm$ 5.28 \\
3  & 97.90 $\pm$ 1.58 & 99.50 $\pm$ 0.21 \\
4  & 88.60 $\pm$ 4.02 & 97.35 $\pm$ 0.94 \\
5  & 69.05 $\pm$ 5.55 & 89.63 $\pm$ 3.67 \\
6  & 85.24 $\pm$ 3.31 & 96.84 $\pm$ 0.70 \\
7  & 84.96 $\pm$ 7.38 & 93.52 $\pm$ 5.60 \\
8  & 84.99 $\pm$ 3.04 & 96.30 $\pm$ 1.20 \\
9  & 87.33 $\pm$ 10.24 & 94.97 $\pm$ 5.97 \\
10 & 93.27 $\pm$ 2.92 & 97.27 $\pm$ 0.93 \\
11 & 95.53 $\pm$ 1.73 & 98.38 $\pm$ 0.23 \\
12 & 93.34 $\pm$ 6.60 & 99.05 $\pm$ 0.35 \\
13 & 86.27 $\pm$ 1.65 & 97.17 $\pm$ 1.23 \\
\hline
\textbf{Mean} & \textbf{87.29 $\pm$ 5.14} & \textbf{96.89 $\pm$ 1.99} \\
\textbf{SD}   & \textbf{8.00 $\pm$ 3.95}  & \textbf{2.74 $\pm$ 2.05} \\
\hline
\end{tabular}
\end{table}

A non-parametric Friedman test was used to compare the nine methods across 13 subjects, using each subject’s mean classification accuracy as the dependent variable. The analysis revealed a significant difference among the classifiers, $\chi^2(8) = 85.78$, $p < 0.001$, indicating that not all models performed equivalently. The corresponding Kendall’s coefficient of concordance ($W = 0.825$) suggested a large effect size, confirming strong agreement in performance ranking across participants. The average ranks were as follows: CCA+BLDA (1.08), Corr+BLDA (2.23), Siamese (multi-class) (3.62), CNN--$K$~bit (Constrained-EMD) (4.46), CNN+$N_{classes}$ (4.77), CNN--$K$~bit (EMD) (5.62), CNN--$K$~bit (Mahalanobis) (6.85), CNN--$K$~bit (Euclidean) (8.15), and Siamese (multiple-classifier) (8.23).

Post-hoc pairwise comparisons were performed using the Wilcoxon signed-rank test with Bonferroni correction ($\alpha_{\text{adj}} = 0.0014$) to control for multiple testing. Significant differences ($p < 0.05$ after correction) were observed between the traditional BLDA-based baselines and most CNN and Siamese models, confirming that the deep learning approaches produced statistically distinct results. In particular, Corr+BLDA and CCA+BLDA yielded significantly lower performance ($p=0.0088$) compared with nearly all CNN variants and the Siamese multiple-classifier model.
 Among the CNN--$K$~bit configurations, the Constrained-EMD and EMD distance metrics did not differ significantly, whereas both were superior to the Euclidean ($p=0.0088$) and Mahalanobis ($p=0.0088$) metrics. The CNN+$N_{classes}$ model performed comparably to the Constrained-EMD variant, showing no statistically significant difference between them.

Overall, the statistical analysis demonstrates that the proposed deep architectures outperform the classical correlation- and CCA-based BLDA methods. Within the deep learning group, the CNN —$K$~bit (Constrained-EMD) and Siamese (multiple-classifier) networks exhibited the most consistent performance across participants. In contrast, the Siamese (multi-class) network achieved significantly lower accuracies ($p=0.0088$) than almost all other methods. These findings confirm that incorporating spatial–temporal feature learning with distance-based decoding yields a more robust and generalizable representation of C-VEP responses.

\subsection{Data Augmentation and classifier combination}
\label{sec:data_aug}

To evaluate the robustness of the proposed deep learning models against temporal variability in EEG signals, a series of controlled data augmentation experiments were conducted. Temporal augmentation was performed by shifting the EEG signal in the time domain by a fixed number of samples to simulate small latency variations in the visual response. Specifically, each trial was shifted by $\pm1$, $\pm2$, $\pm4$, and $\pm8$ time points, and new training and testing sets were generated for each shift condition. Three augmentation strategies were examined: (i) Train Augmentation (TA), where only the training data were augmented; (ii) Test Combination (TC), where the score from different shifted inputs are averaged; and (iii) Train Augmentation and Test Combination (TA\&TC), where the training set was augmented and the scores were combined. The baseline condition (NA) corresponds to models trained and evaluated without temporal shifting.

\begin{figure*}
    \centering
\begin{tabular}{cc}
\includegraphics[width=0.45\linewidth]{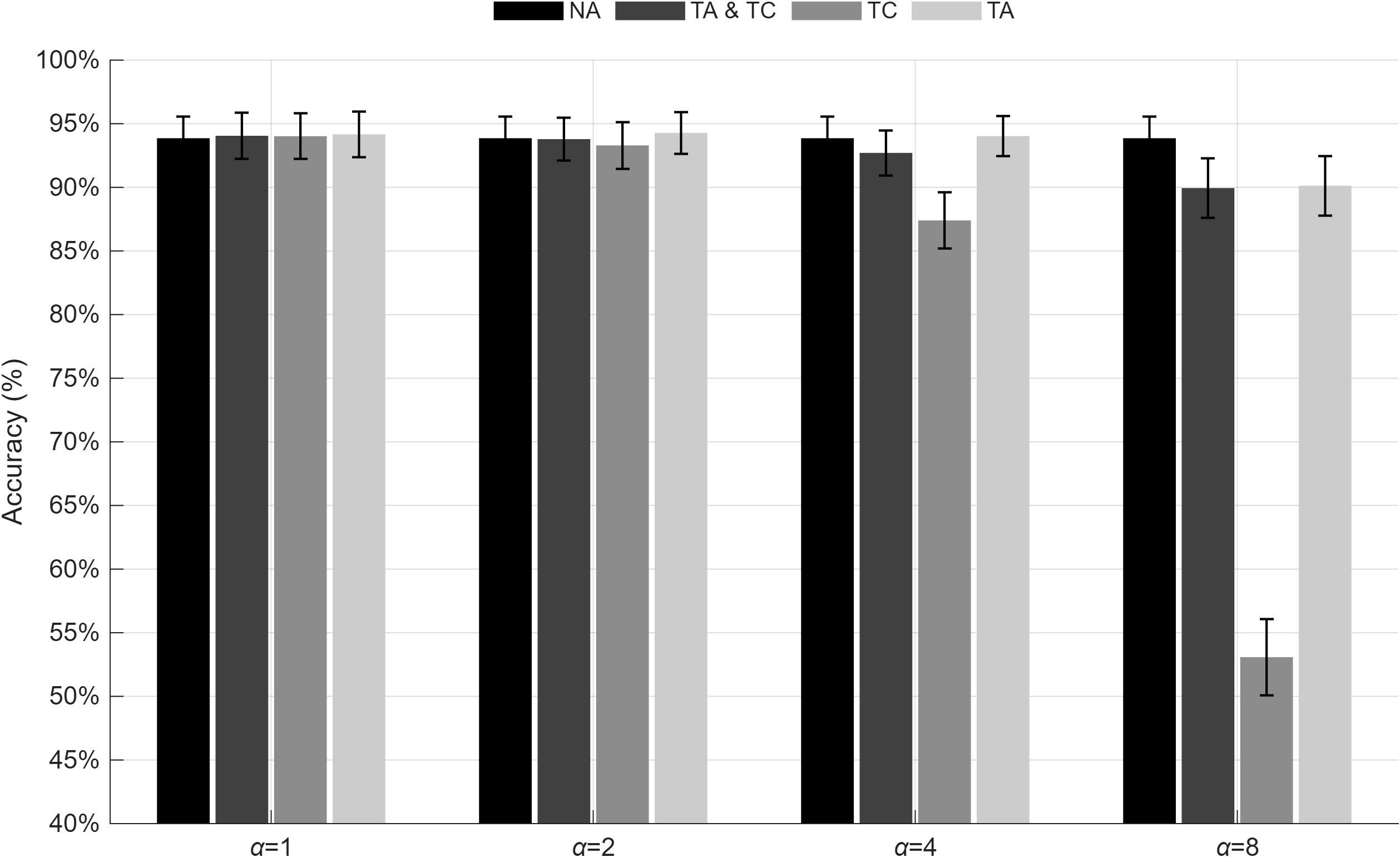} & 
\includegraphics[width=0.45\linewidth]{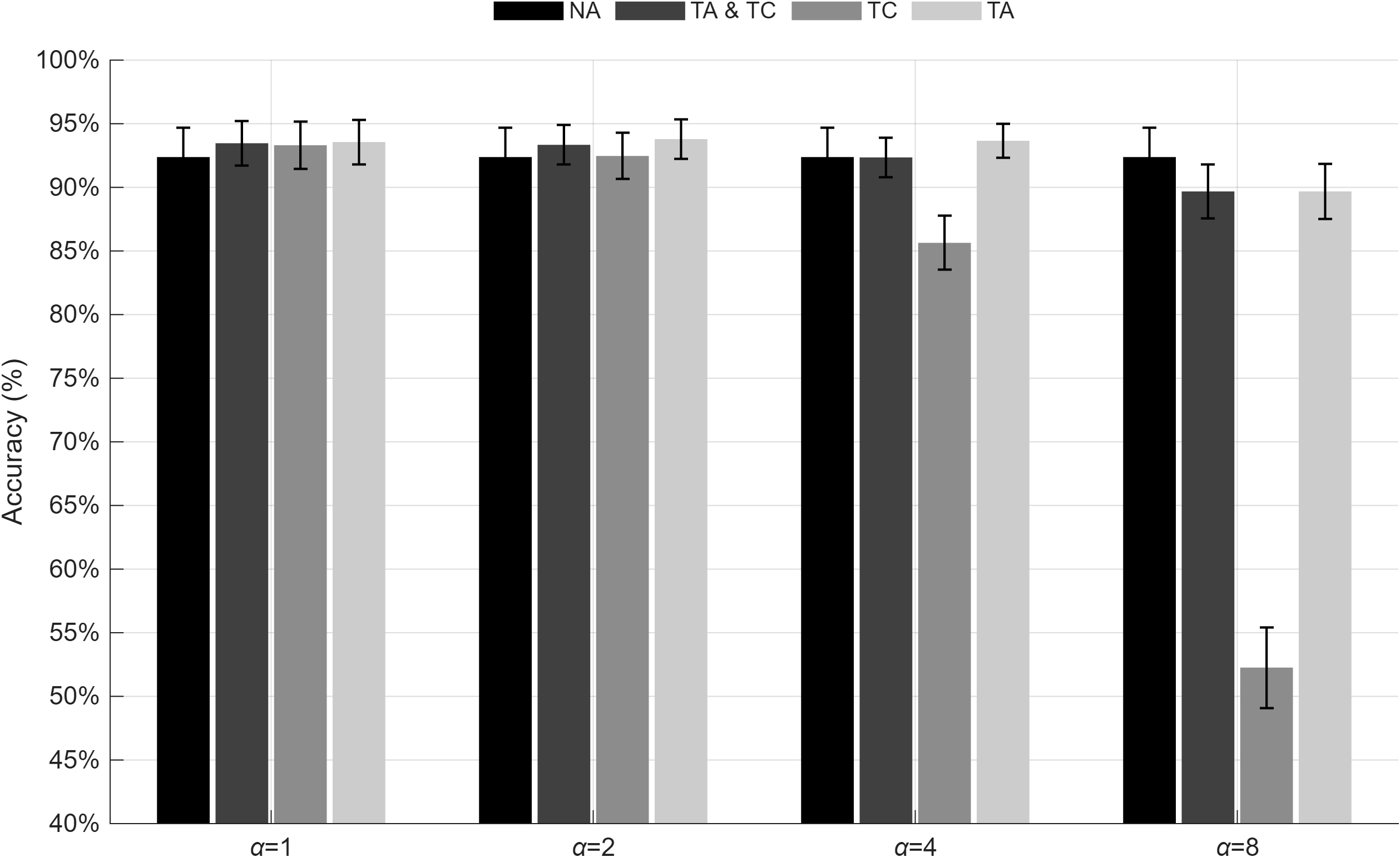} \\
CNN~$K$-bit (Euclidean distance) & CNN~$K$-bit (Mahalanobis distance) \\
\includegraphics[width=0.45\linewidth]{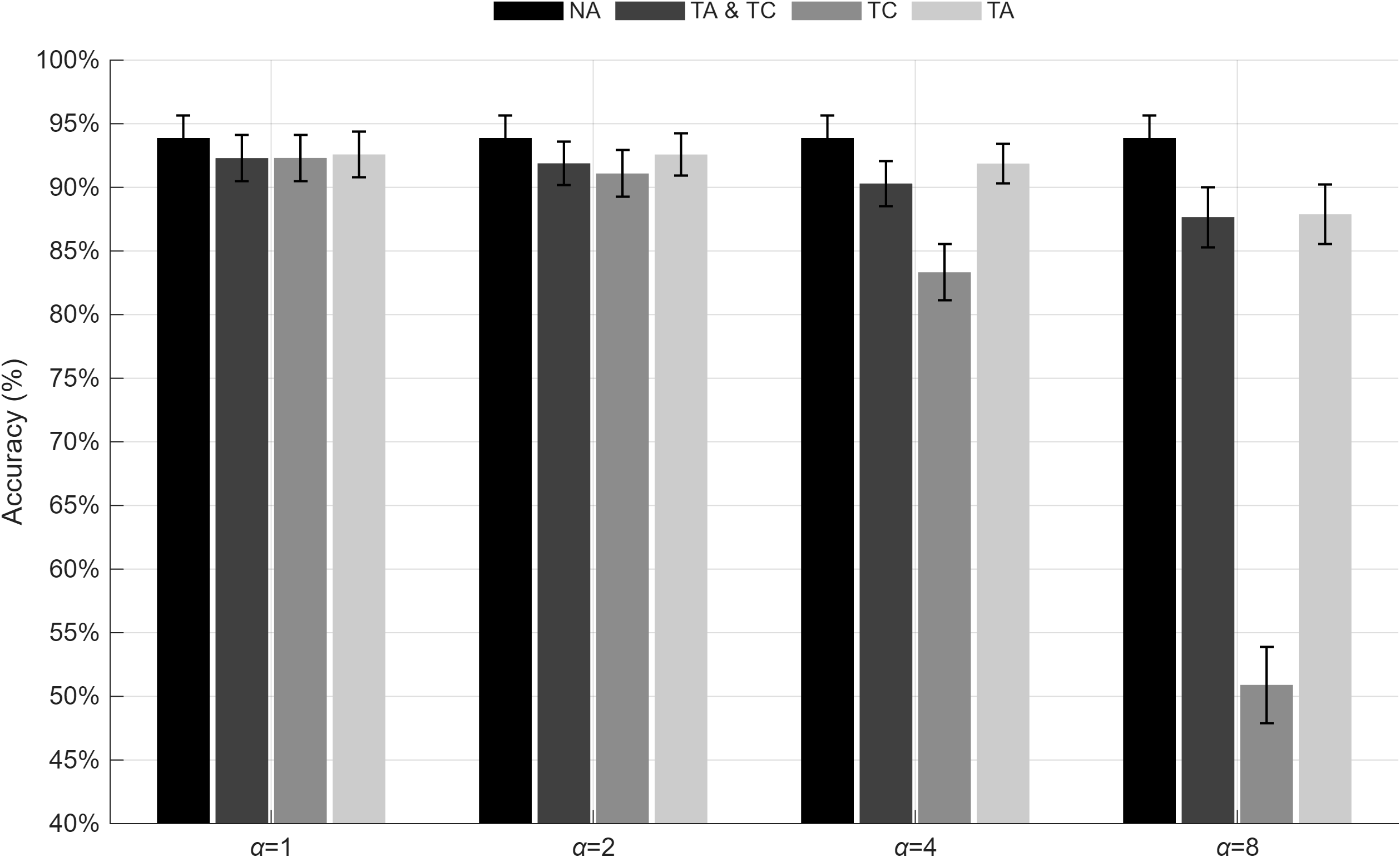} & 
\includegraphics[width=0.45\linewidth]{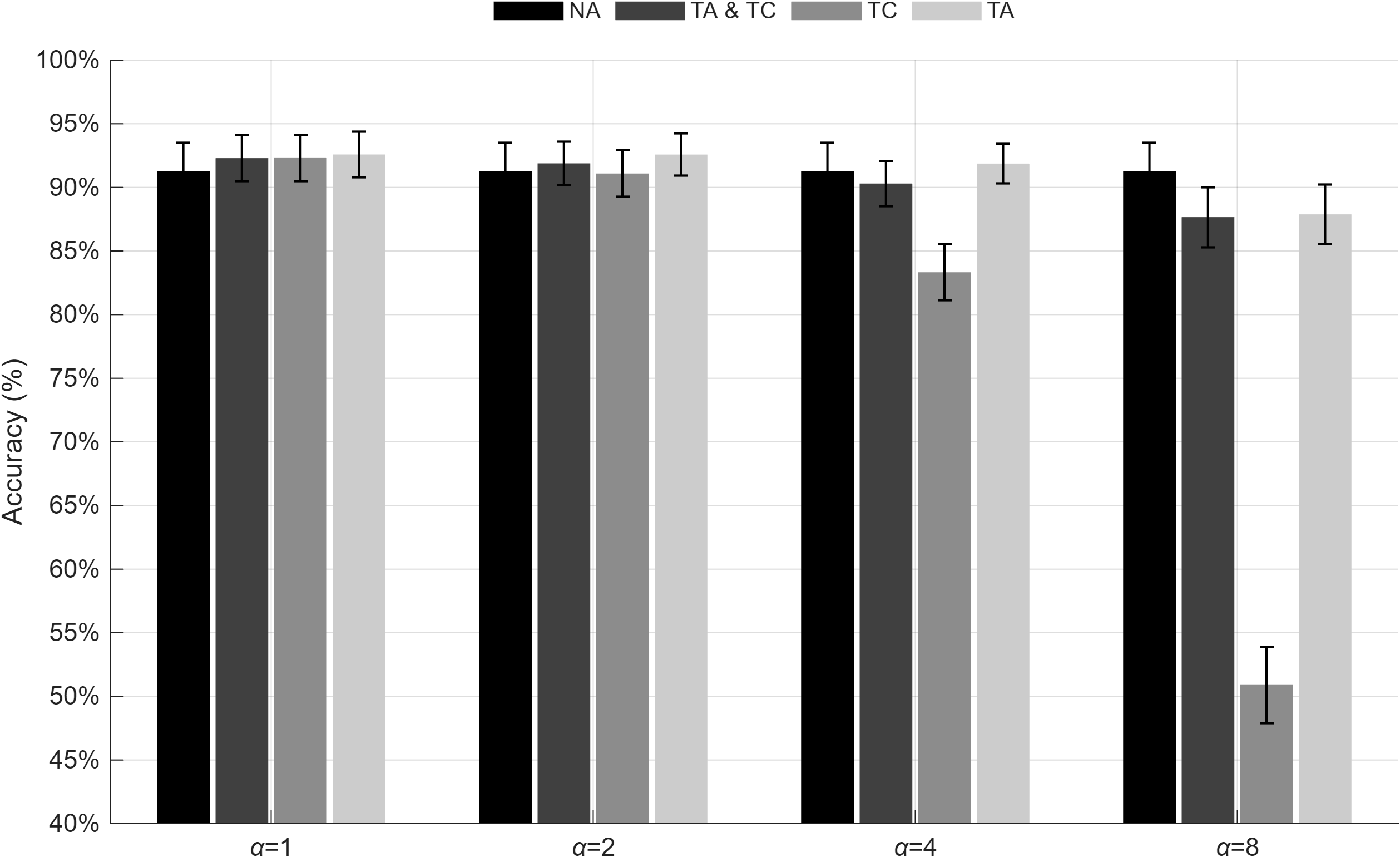} \\
CNN~$K$-bit (Earth Mover’s Distance) & CNN~$K$-bit (Constrained Earth Mover’s Distance)
\end{tabular}
 \caption{Performance comparison of CNN~$K$-bit models under different distance metrics and temporal augmentation conditions.}
    \label{fig:data_aug_cnn63_all}
\end{figure*}

\begin{figure}
    \centering
    \includegraphics[width=0.5\linewidth]{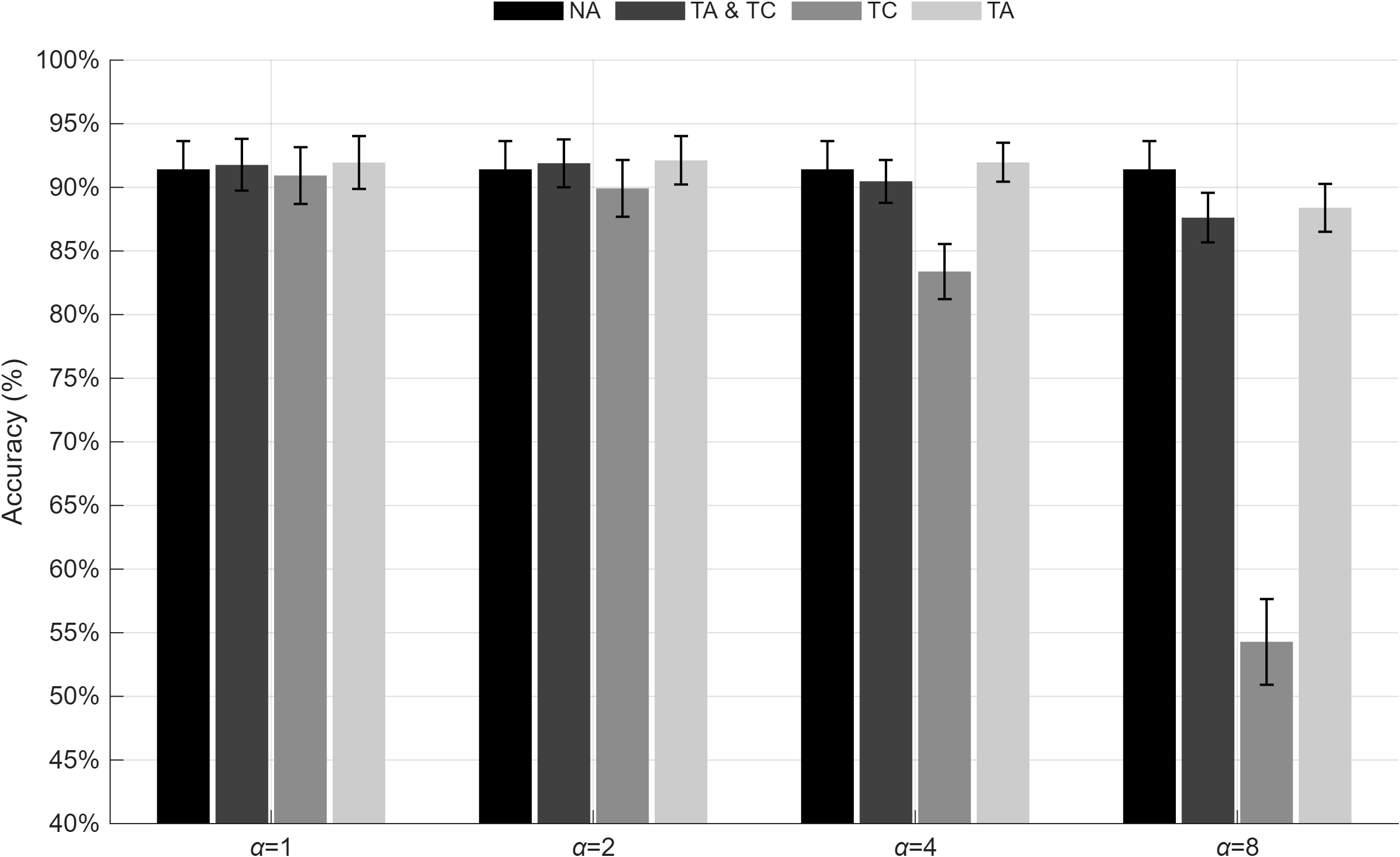}
    \caption{Performance of the CNN~+$N_{classes}$ model under various temporal augmentation conditions and shift magnitudes.}
    \label{fig:data_aug_cnn6class}
\end{figure}

\begin{figure*}
    \centering
    \begin{subfigure}[t]{0.48\textwidth}
        \centering
        \includegraphics[width=\linewidth]{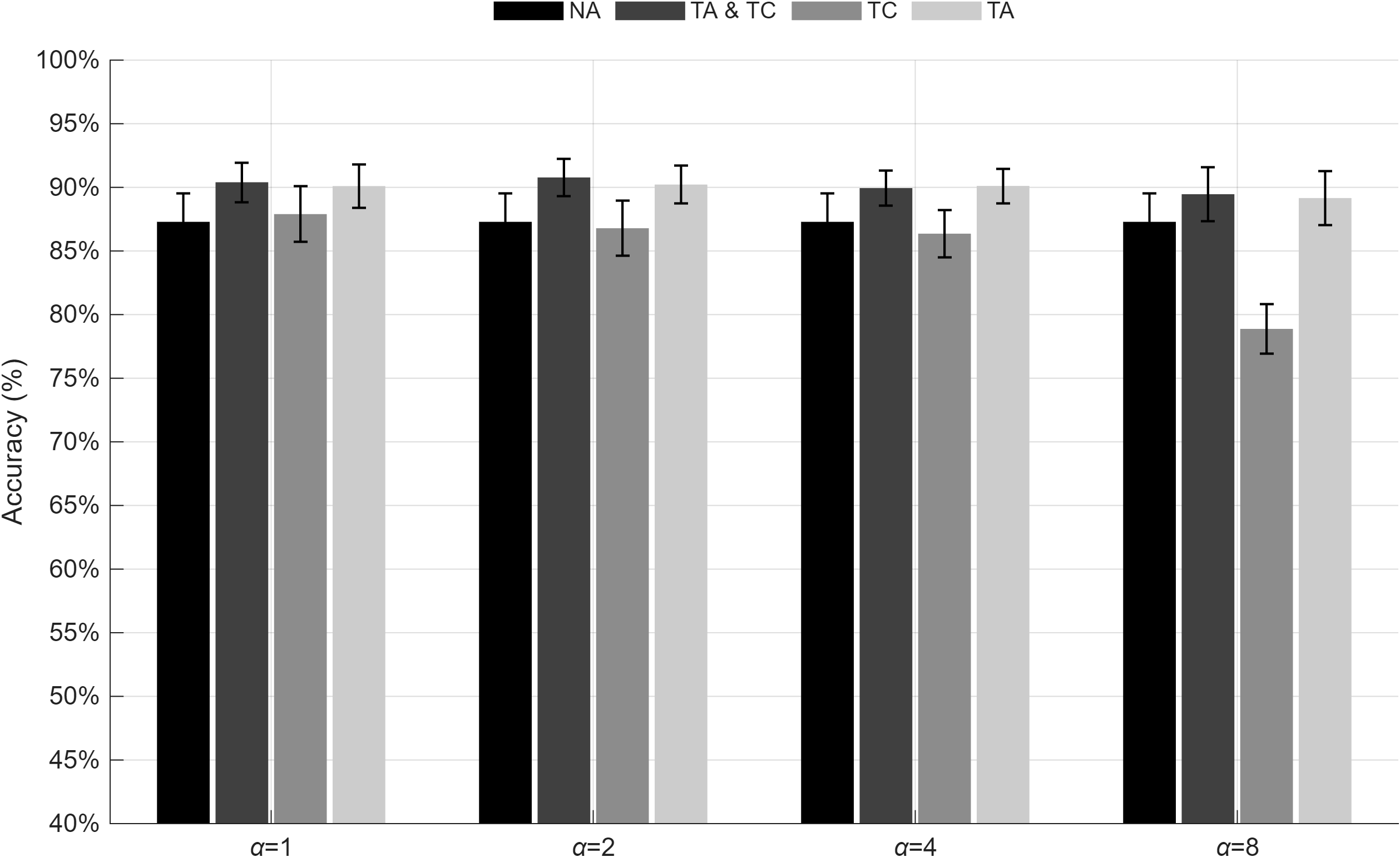}
        \caption{Single multi-class Siamese.}
        \label{fig:data_aug_siam_binary}
    \end{subfigure}
    \hfill
    \begin{subfigure}[t]{0.48\textwidth}
        \centering
        \includegraphics[width=\linewidth]{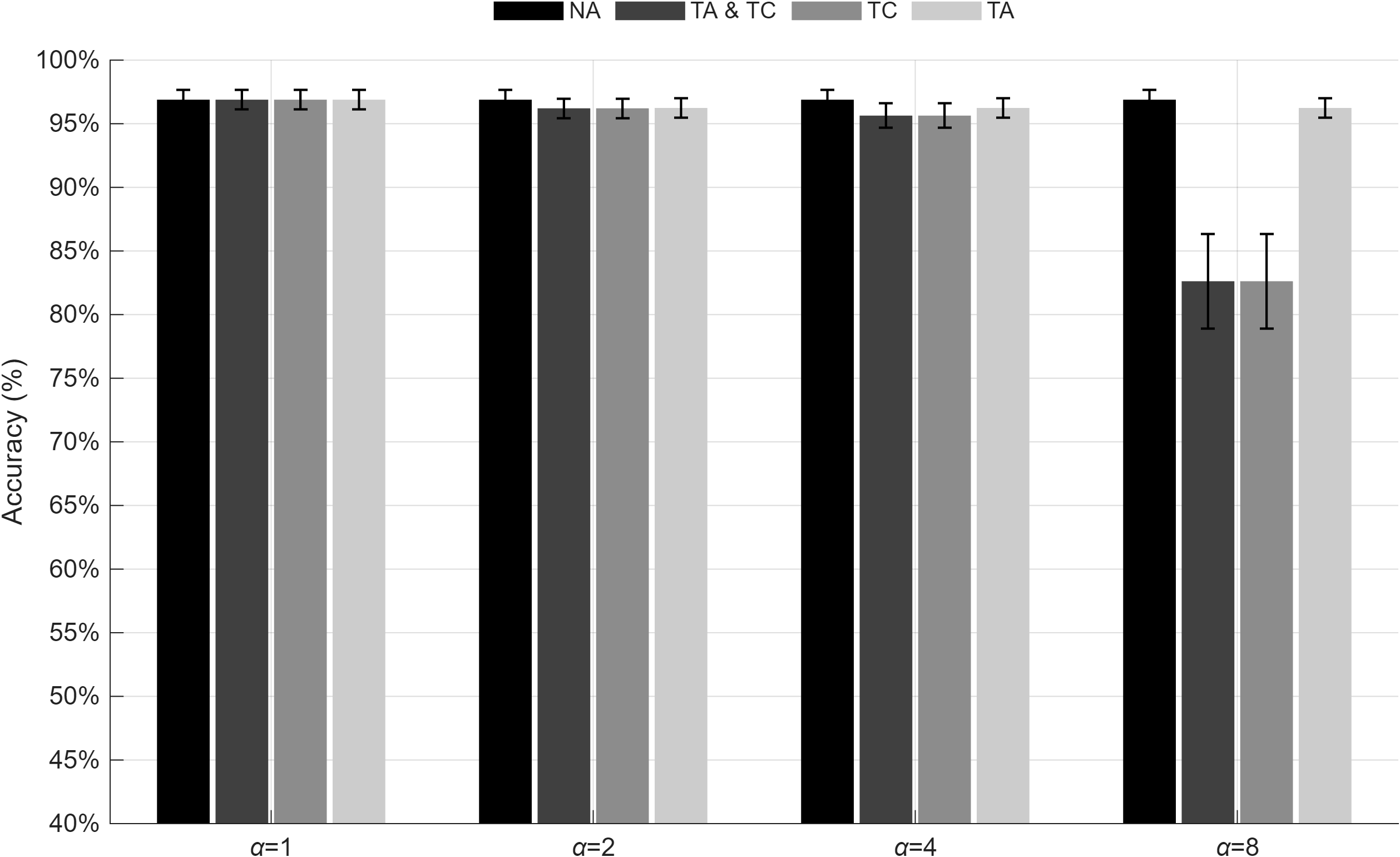}
        \caption{Multiple-classifier binary Siamese.}
        \label{fig:data_aug_siam_multi}
    \end{subfigure}

    \caption{Performance of Siamese network models under different temporal augmentation conditions and shift magnitudes. 
    Comparison between the Single multi-class Siamese  and Multiple-classifier binary Siamese architectures.}
    \label{fig:data_aug_siamese_all}
\end{figure*}

This augmentation framework was applied consistently across all deep learning architectures. As illustrated in Figs.~\ref{fig:data_aug_cnn63_all}, ~\ref{fig:data_aug_cnn6class},~\ref{fig:data_aug_siam_binary}, and ~\ref{fig:data_aug_siam_multi},
temporal augmentation up to $\alpha = 4$ samples improved overall generalization and reduced performance variability across sessions. In contrast, excessive temporal shifting ($\alpha = 8$) led to a gradual decline in accuracy for most methods, particularly when augmentation was applied only to the testing data (TC).

Across all models, the first column of each accuracy matrix represents the baseline (NA) condition, followed by TA\&TC, TC, and TA configurations. Under moderate temporal shifts ($\alpha \leq 4$), most models demonstrated stable or improved performance compared to the non-augmented baseline. For the CNN~$K$-bit models, accuracies remained above 92\% for all shift levels up to $\alpha = 4$, with peak values observed for the Euclidean ($94.17 \pm 6.44\%$) and Mahalanobis ($93.79 \pm 5.59\%$) variants under TA. The EMD and Constrained-EMD models achieved comparable stability, maintaining approximately $92.6 \pm 5$–$6\%$ accuracy at $\alpha = 2$–$4$, before declining to around $87$–$88\%$ at $\alpha = 8$ when TC was applied. The CNN+$N_{classes}$ exhibited a similar pattern, improving from $91.42 \pm 7.93\%$ (NA) to $92.12 \pm 6.84\%$ under TA at $\alpha = 2$, but dropping sharply to $54.29 \pm 12.15\%$ at $\alpha = 8$ for TC-only conditions. The Siamese networks were notably resilient: the multi-class model increased from $87.29 \pm 8.00\%$ (NA) to $90.22 \pm 5.34\%$ under TA at $\alpha = 2$, while the multiple-classifier variant maintained consistently high performance near $96.2 \pm 2.7\%$ up to $\alpha = 4$, followed by a marked decline to $82.62 \pm 13.38\%$ at $\alpha = 8$ for TC and TA\&TC. Overall, augmentation of the training data (TA and TA\&TC) consistently improved temporal robustness, whereas applying shifts solely to the test set (TC) reduced accuracy and increased variance across subjects.

\section{Discussion}
\label{section:discussion}

We investigated the performance of multiple deep learning and classical models for single-trial decoding of C-VEPs. The approaches varied in their principles and levels, with discriminant approaches that classify input signals directly or reproduce signals for use in a density-based solution as they are compared with templates at a later stage. In deep learning approaches, layers dedicated to high-level feature extraction were common, and we observed substantial differences in how the later stages of the network were used.  

The experiments demonstrated that convolutional and Siamese network architectures significantly outperform traditional BLDA and CCA-based classifiers. Among all approaches, the CNN~$K$-bit models and Siamese networks achieved superior accuracy, exceeding 90\% across subjects, while the classical Corr+BLDA and CCA+BLDA baselines achieved 81.64~$\pm$~14.02\% and 62.59~$\pm$~16.57\%, respectively. The improvements achieved by deep learning methods highlight their ability to learn complex temporal–spatial dependencies from raw EEG signals, capturing subtle nonlinearities that conventional linear models cannot represent.

The comparative analysis among CNN distance metrics revealed that reconstruction-based models decoded the $K$-bit m-sequence most effectively when coupled with perceptually consistent measures such as the EMD and its constrained variant. Both CNN~$K$-EMD (93.88~$\pm$~6.38\%) and CNN~$K$-C-EMD (91.30~$\pm$~7.99\%) exhibited greater robustness to inter-session variability compared to Euclidean (93.86~$\pm$~6.16\%) and Mahalanobis (92.38~$\pm$~8.34\%) metrics. This finding suggests that EMD-based decoding better preserves temporal structure and accounts for small phase distortions inherent in EEG recordings. Similarly, the CNN~+$N_{classes}$ achieved 91.42~$\pm$~7.93\% accuracy, indicating that sequence-level learning can be effectively replaced by categorical learning without substantial performance degradation. The Siamese architectures further demonstrated the capacity of feature-space similarity learning for EEG-based classification, with the multi-class variant (87.29~$\pm$~8.00\%) providing a compact yet effective representation and the multiple-classifier model achieving the highest overall accuracy of 96.89~$\pm$~2.74\%.

Temporal data augmentation analysis revealed that moderate temporal shifting enhanced model generalization by simulating variations in physiological response latency. Augmentation up to $\alpha=4$ samples improved stability for both CNN and Siamese models, particularly under TA and TA\&TC configurations, while larger shifts ($\alpha=8$) caused performance degradation, especially in the TC-only setup. For instance, the CNN~$K$-bit (Euclidean) model maintained $94.17~\pm~6.44\%$ under TA at $\alpha=1$, whereas accuracy dropped to $53.08~\pm~10.81\%$ at $\alpha=8$ for TC. Similarly, the Siamese multi-class model improved from 87.29~$\pm~8.00\%$ (NA) to 90.22~$\pm~5.34\%$ (TA, $\alpha=2$), and the Siamese multiple-classifier variant remained nearly invariant ($\approx96.2~\pm~2.7\%$) for $\alpha \leq 4$, followed by a sharp decline ($82.62~\pm~13.38\%$) at $\alpha=8$. These results confirm that augmenting the training data with small temporal perturbations enhances temporal robustness. In contrast, excessive misalignment between training and testing signals introduces inconsistent phase information, which is detrimental to decoding accuracy.

Compared with prior EEG-based C-VEP and SSVEP decoding studies, which typically report 70–85\% accuracy using correlation or canonical methods, the proposed deep learning architectures demonstrate state-of-the-art single-trial decoding performance. The use of m-sequence reconstruction and distance-based similarity learning provides a novel framework that integrates both temporal fidelity and representational flexibility. Moreover, the inclusion of augmentation and inter-session validation establishes the practical relevance of these models for robust, session-independent BCI design.

Although the results are promising, several limitations should be acknowledged. The number of electrodes was limited to eight sites in the occipital and parietal regions, potentially limiting spatial resolution. The models were trained and evaluated within-subject using leave-one-session-out cross-validation, and cross-subject generalization remains to be examined. Future work will aim to extend these findings toward subject-independent training and closed-loop real-time C-VEP BCIs. The integration of multimodal signals, such as electrooculography (EOG), may further enhance decoding reliability and usability in practical neurotechnology systems.

\section{Conclusion}
\label{section:conclusion}

We proposed and compared multiple deep learning approaches for EEG-based code-modulated visual evoked potential (C-VEP) decoding, including convolutional neural networks (CNNs) with both class-level and sequence-level outputs combined with different distance metrics, as well as Siamese network architectures. The results demonstrated that the proposed deep learning models significantly outperform traditional correlation-based methods, with the CNN~$K$-bit models and Siamese networks achieving superior accuracy. Among all models, the multiple-classifier Siamese network achieved the highest overall performance with an accuracy of 96.89\%, followed by distance-based CNN variants that attained accuracies in the range of 91–94\%. In particular, distance-based decoding using Earth Mover’s Distance (EMD) and constrained EMD improved temporal alignment and model generalization, while temporal data augmentation, even with small shifts, enhanced stability against latency variations. These findings highlight the advantages of end-to-end representation learning in capturing the complex temporal–spatial structure of EEG signals, representing a significant step toward practical, reliable, and adaptive non-invasive C-VEP brain–computer interfaces.

\subsection{Acknowledgment}

This work has been supported by the NIH-R15 NS118581 grant.
The authors would like to thank all participants who contributed to the EEG data collection.

\bibliographystyle{unsrtnat} 
\bibliography{references}

\end{document}